\documentclass[letterpaper, 10 pt, journal]{IEEEtran}
\IEEEoverridecommandlockouts    
\usepackage{hyperref}

\hypersetup{
    colorlinks=true,
    linkcolor=blue,
    filecolor=magenta,      
    urlcolor=blue,
}

\newcommand{\tabincell}[2]{\begin{tabular}{@{}#1@{}}#2\end{tabular}}

\usepackage{graphics}           
\usepackage{times}              
\usepackage{amsmath}            
\usepackage{amssymb}            
\usepackage{graphicx}
\usepackage{algorithm}
\usepackage[noend]{algpseudocode}
\usepackage{booktabs}
\usepackage{color}
\definecolor{instructioncolor}{rgb}{.5,.5,.5}

\usepackage[font=small]{caption}

\def\secref#1{Sec.~\ref{#1}}
\def\figref#1{Fig.~\ref{#1}}
\def\tabref#1{Tab.~\ref{#1}}
\def\eqref#1{Eq.~(\ref{#1})}

\makeatletter
\usepackage{xspace}
\DeclareRobustCommand\onedot{\futurelet\@let@token\@onedot}
\def\@onedot{\ifx\@let@token.\else.\null\fi\xspace}

\def\etal{{et al}\onedot}
\makeatother

\usepackage{array}
\newcolumntype{L}[1]{>{\raggedright\let\newline\\\arraybackslash\hspace{0pt}}m{#1}}
\newcolumntype{C}[1]{>{\centering\let\newline\\\arraybackslash\hspace{0pt}}m{#1}}
\newcolumntype{R}[1]{>{\raggedleft\let\newline\\\arraybackslash\hspace{0pt}}m{#1}}

\def\argmax{\mathop{\rm argmax}}

\newcommand{\RR}{\mathbb{R}}

\usepackage{subfigure}
\usepackage{multirow}
\usepackage{rotating}  

\title{PCPNet: An Efficient and Semantic-Enhanced Transformer Network for Point Cloud Prediction}

\author{Zhen Luo ~\and Junyi Ma ~\and Zijie Zhou ~\and Guangming Xiong$^*$ 
  \thanks{
  Z. Luo, J. Ma, Z. Zhou and Guangming Xiong are with Beijing Institute of Technology.
  }
  \thanks{$^*$corresponding author email: xiongguangming@bit.edu.cn}
}

\begin{document}
\maketitle

\IEEEpeerreviewmaketitle
\thispagestyle{empty}
\pagestyle{empty}

\begin{abstract}

The ability to predict future structure features of environments based on past perception information is extremely needed by autonomous vehicles, which helps to make the following decision-making and path planning more reasonable. Recently, point cloud prediction (PCP) is utilized to predict and describe future environmental structures by the point cloud form. In this letter, we propose a novel efficient Transformer-based network to predict the future LiDAR point clouds exploiting the past point cloud sequences. We also design a semantic auxiliary training strategy to make the predicted LiDAR point cloud sequence semantically similar to the ground truth and thus improves the significance of the deployment for more tasks in real-vehicle applications. Our approach is completely self-supervised, which means it does not require any manual labeling and has a solid generalization ability toward different environments. The experimental results show that our method outperforms the state-of-the-art PCP methods on the prediction results and semantic similarity, and has a good real-time performance. Our open-source code and pre-trained models are available at \url{https://github.com/Blurryface0814/PCPNet}.\\

\end{abstract}
\begin{IEEEkeywords}
point cloud prediction, semantic auxiliary training, self-supervised learning.
\end{IEEEkeywords}

\section{Introduction}
\label{sec:intro}

Sequential 3D point clouds can be used to accomplish multiple complex tasks for autonomous vehicles such as simultaneous localization and mapping (SLAM)~\cite{shan2018lego, shan2020lio}, place recognition~\cite{ma2022ral, ma2022seqot}, object detection~\cite{zhou2018voxelnet, chen2022mppnet}, and semantic segmentation~\cite{milioto2019rangenet++, liu2019meteornet}. Recently, exploiting the past 3D point cloud sequence to predict the future 3D point cloud sequence, also known as point cloud prediction (PCP), has attracted more attention in the field of point cloud processing~\cite{fan2019pointrnn, zhang2021cloudlstm, deng2020temporal, lu2021monet, weng2021inverting, weng2022s2net, mersch2022self}. The predicted point clouds can be directly utilized by the existing point cloud processing methods to further realize future object detection and semantic segmentation~\cite{lu2021monet, weng2021inverting}. Therefore, PCP methods deployed on autonomous vehicles can greatly improve the ability to perceive future driving conditions, thus leading to more reasonable decision-making and path planning in the following driving strategy. The point cloud sequence used by PCP methods is ordered in the time dimension but is unordered in the space dimension within each observation. Compared to the existing video prediction methods~\cite{kwon2019predicting, jin2020exploring, wu2021motionrnn}, 3D point cloud prediction needs to use LiDAR data with both ordered and unordered features to solve the sequence-to-sequence problem. In addition, the sparsity of LiDAR point clouds further increases the difficulty of this task since it is hard to capture the current structure information from discrete laser points and then predict the future point clouds.

\begin{figure}
  \centering
  \includegraphics[width=1\linewidth]{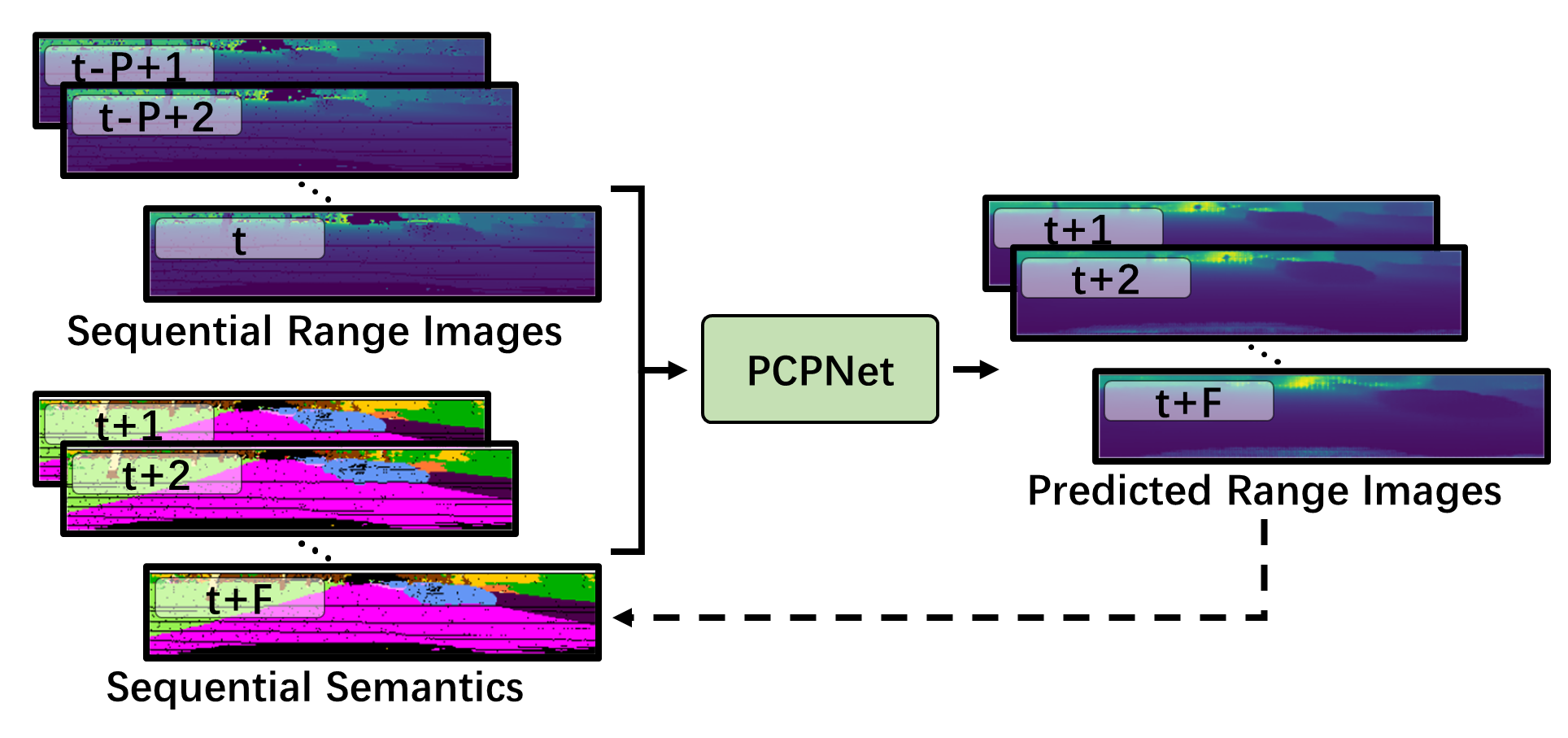}
  \caption{PCPNet predicts future $F$ range images based on the given past $P$ sequential range images. The semantic information in the sequential range images is extracted for training, making the outputs of PCPNet closer to the ground truth in semantics.}
  \label{fig:motivation}
  \vspace{-0.5cm}
\end{figure}

In this paper, we propose an efficient point cloud prediction network named PCPNet, as shown in~\figref{fig:motivation}. PCPNet exploits Transformer to capture the inner correlation within each sequential LiDAR observations. Compared with the previous works that directly apply Transformer to point cloud features~\cite{guo2021pct, fan2021point}, we first convert the point cloud to the range image, and then compress it along the height and width dimension respectively to generate the sentence-like features for the following Transformer. In contrast to sparse point clouds, the orderly arrangement of pixels in the range image makes the self-attention mechanism of Transformer works better. To further make the predicted point clouds semantically similar to ground truth, we also design an auxiliary training strategy that utilizes the semantics of the predicted point clouds to enhance the prediction performance. The devised semantic-based loss function helps to increase the semantic similarities between the predicted point clouds and the ground truth point clouds, leading to more significant information for other algorithms deployed on autonomous vehicles such as high-level point cloud processing and trajectory prediction. To the best of our knowledge, this is the first work that introduces Transformer and semantic enhancement into point cloud prediction. In addition, the overall system of PCPNet is entirely self-supervised since the input point clouds and ground truth ones are both from real-time sequential LiDAR observations. Note that we only use a pre-trained semantic segmentation network to generate semantic labels online for auxiliary training. This makes our proposed approach collect training data in any driving conditions automatically without intensive manual labeling.

To validate the good prediction performance and solid generalization ability of our proposed method, we conduct several experiments on publicly available datasets to compare our approach with the existing state-of-the-art PCP baselines. We also provide an ablation study to demonstrate that our proposed network structure is reasonable and efficient. Besides, we design an experiment to validate the effectiveness of the proposed semantic auxiliary training strategy.

Our contributions can be summarized as follows:

\begin{itemize}
\item[$\bullet$] A Transformer-based neural network with encoder-decoder architecture named PCPNet is proposed, which achieves state-of-the-art performance on point cloud prediction.
\item[$\bullet$] A lightweight semantic segmentation network is integrated to provide semantic labels for auxiliary training, which increases the semantic accuracy of the predicted point clouds.
\item[$\bullet$] The overall system is trained in a self-supervised manner to improve the generalization ability and avoid labor-intensive labeling in real-vehicle applications.
\end{itemize} 

\section{Related Work}
\label{sec:related}

To capture future features of the environments using past information, video prediction~\cite{kwon2019predicting, jin2020exploring, wu2021motionrnn} has been well investigated in the field of computer vision. In contrast, point cloud prediction is a novel topic in robotics and only a few studies have been working on it. PointRNN proposed by Fan \etal~\cite{fan2019pointrnn} adopts a point-based spatiotemporally-local correlation to aggregate point features and states according to point coordinates to model and predict point cloud sequences. Zhang \etal~\cite{zhang2021cloudlstm} design a dynamic convolution operator which allows performing convolution operations directly over point clouds. Deng \etal~\cite{deng2020temporal} propose a method based on FlowNet3D and Dynamic Graph CNN to predict future LiDAR frames. They use the point-based feature extractor and furthest point sampling to generate end-to-end architectures that operate directly on point clouds. Lu \etal~\cite{lu2021monet} propose a motion-based neural network named MoNet which integrates the motion features between two consecutive point clouds into the RNN prediction pipeline. Different from the above-mentioned methods that only use raw point clouds as input, several existing methods also utilize sequential range images as input to further improve the operation efficiency. SPFNet by Weng \etal~\cite{weng2021inverting} uses a shared encoder and an LSTM to extract features from every past scan and model temporal dynamics respectively. These features are then fed to a shared decoder to predict future scans. Based on SPFNet, Weng \etal~\cite{weng2022s2net} propose a stochastic SPFNet named S2Net, which can sample sequences of latent variables to tackle future uncertainty. Mersch \etal~\cite{mersch2022self} propose a 3D range-image-based method for predicting future point clouds. They project past point cloud sequences as 2D range images and then concatenate them to generate 3D tensors. A 3D spatio-temporal CNN with encoder-decoder architecture is designed to predict future point clouds based on these tensors.

The main challenge in point cloud prediction is how to extract significant spatio-temporal features from the sparse and disordered point cloud sequences. The existing trajectory tracking and moving object segmentation approaches have designed diverse networks to extract spatio-temporal features in a learning-based manner. Alahi \etal~\cite{alahi2016social} use LSTM to extract spatio-temporal information to predict pedestrian trajectories. Huang \etal~\cite{huang2019stgat} model the interaction of tracking objects at each time step by a graph neural network. Compared to the RNN-based methods, more and more CNN-based methods have been proposed to achieve state-of-the-art performance in recent years. For example, Chen \etal~\cite{chen2021moving} design an encoder-decoder network with combined residual images as input to improve the performance of moving object segmentation. Sun \etal~\cite{sun2022efficient} use a range-image-based dual-branch structure to process spatial and temporal information respectively, and then combine them with motion-guided attention modules. 
In recent years, Transformer \cite{vaswani2017attention} became a more trendy way to extract spatial features within one single LiDAR scan, but few works use it to extract temporal features from sequential LiDAR observations. Fan \etal~\cite{fan2021point} apply a point-based Transformer for spatio-temporal modeling of small-scale point cloud video. Ma \etal~\cite{ma2022seqot} utilize sequential range images as network input and extract significant features using Transformer to recognize previously seen places.

In this paper, we propose PCPNet which first uses Transformer to aggregate spatial and temporal information from sequential range images to forecast point clouds. Besides, compared to the existing PCP methods which only focus on low-level structure features, PCPNet also uses high-level semantic information for auxiliary training to further improve the prediction performance.

%

\begin{figure*}
  \centering
  \includegraphics[width=0.9\linewidth]{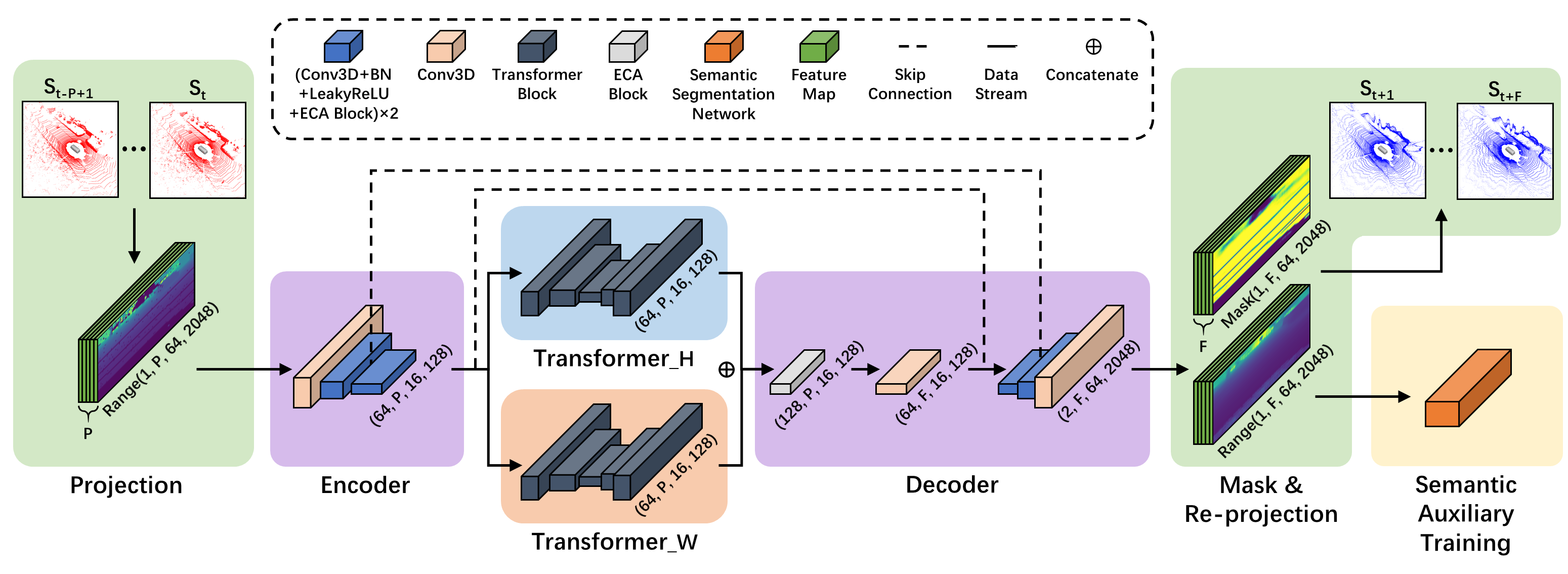}
  \caption{The overall architecture of PCPNet. The numbers in the bracket below each cube represent the size of the feature map output from this layer. The input range images are first downsampled and compressed along the height and width dimensions respectively to generate the sentence-like features for the following Transformer blocks. The features are then combined and upsampled to the predicted range images and mask images. Semantic auxiliary training is used to enhance the practical value of point cloud prediction.}
  \label{fig:overall}
  \vspace{-0.5cm}
\end{figure*}

\section{Our Approach}
\label{sec:method}

\subsection{Overall Architecture}
\label{sec:overall_arch}

A point cloud sequence with the length of $T$ can be described as $S = \lbrace S_{1},S_{2},...,S_{t},...,S_{T}\rbrace$, where $S_{t}$ represents the point cloud frame at time $t$. Further, $S_{t} = \lbrace p^{t}_{i} ~\vert ~ i = 1,2,...,N_{t} \rbrace$ contains $N_{t}$ unordered points, where $p^{t}_{i} \in \RR^3$ represents a three-dimensional coordinate vector. Point cloud prediction is to forecast the future point cloud sequence $\lbrace S_{t+1},S_{t+2},...,S_{t+F}\rbrace$ with the length of $F$ based on the given past point cloud sequence $\lbrace S_{t-P+1},S_{t-P+2},...,S_{t}\rbrace$ with the length of $P$.

To solve this problem, we propose a Transformer-based network named PCPNet. Considering the advantages of Transformer in processing sequential data, we first convert 3D LiDAR point clouds to 2D range images by spherical projection to obtain dense and ordered input data. Specifically, we project a laser point $p_{i}=(x,y,z) \in \RR^3$ to spherical coordinates and finally to image coordinates $(u,v) \in \RR^2$ via a mapping $\Pi:~\RR^3 \mapsto \RR^2$
\begin{align}
  \left( \begin{array}{c} u \vspace{0.5em}\\ v \end{array}\right) & = \left(\begin{array}{cc} \frac{1}{2}\left[1-\arctan(y, x) \pi^{-1}\right] w   \vspace{0.5em} \\
      \left[1 - \left(\arcsin(z r^{-1}) + \mathrm{f}_{\mathrm{up}}\right) \mathrm{f}^{-1}\right] h\end{array} \right), \label{eq:projection}
\end{align}
where $(h,w)$ represents the height and width of the range image, $\mathrm{f}=\mathrm{f}_{\mathrm{up}}+\mathrm{f}_{\mathrm{down}}$ is the vertical field-of-view of the sensor, and $r=\Vert p_{i} \Vert_{2}$ represents the range of each laser point, which is the distance to the sensor origin. We assume that the point clouds are located in the current local coordinate frame of the LiDAR sensor. If no laser point exists for a corresponding pixel, we set $r=0$, and if a pixel corresponds to more than one laser point, the nearest one is retained. In addition, the re-projection $\Pi^{\prime}:~\RR^2 \mapsto \RR^3$ of a pixel on a range image can be described as
\begin{align}
  \left( \begin{array}{c} x \vspace{0.2em}\\ y \vspace{0.2em}\\ z \end{array}\right) & = \left(\begin{array}{cc} r \cos(\theta) \cos(\gamma)  \vspace{0.2em} \\ 
  r \cos(\theta) \sin(\gamma)  \vspace{0.2em} \\ r \sin(\theta) \end{array} \right), \label{eq:re-projection}
\end{align}
where $\theta=\arctan(y,x)$ represents the pitch angle and $\gamma=\arcsin(z r^{-1})$ represents the yaw angle.

The overall architecture of PCPNet is shown in~\figref{fig:overall}. Following the operation of Mersch \etal~\cite{mersch2022self}, we concatenate $P$ range images with height $H_{p}$ and width $W_{p}$ along the time dimension to get a 4D input tensor $\left( C_{p},P,H_{p},W_{p}\right) $, where $C_{p}$ is the number of channels. In this work, $H_{p}$ is set to $64$ and $W_{p}$ is set to $2048$. Since we only use range values as input, $C_{p}=1$ holds. The encoder uses the combination of 3D convolution, 3D batch normalization, LeakyReLU, and ECA block~\cite{2020ECA} to compress the input tensor in the height and width dimensions while maintaining the size of the time dimension. ECA is a channel attention mechanism that we use to enable the network to automatically adjust the weight of each channel. The output tensor from the encoder is then fed into both Transformer\_H block and Transformer\_W block for self-attention. The outputs of the two branches are combined by an ECA block and a 3D convolution, ultimately being fed to the decoder that mirrors the encoder. The final output tensor size of the network is $\left( 2,F,H_{f},W_{f}\right) $, including a range image sequence and a mask image sequence both with size $\left( 1,F,H_{f},W_{f}\right) $. As done by ~\cite{weng2021inverting, mersch2022self}, the mask image sequence contains a probability for each range image pixel to be a valid point for re-projection. We combine and re-project the two image sequences to 3D coordinate system to obtain the final predicted point cloud sequence using \eqref{eq:re-projection}. Note that we use circular padding to maintain spatial consistency on the horizontal borders of the range images in PCPNet. We also use the skip connection to maintain the details of the prediction, which is also shown in~\figref{fig:overall}.
\begin{figure*}
  \centering
  \includegraphics[width=0.9\linewidth]{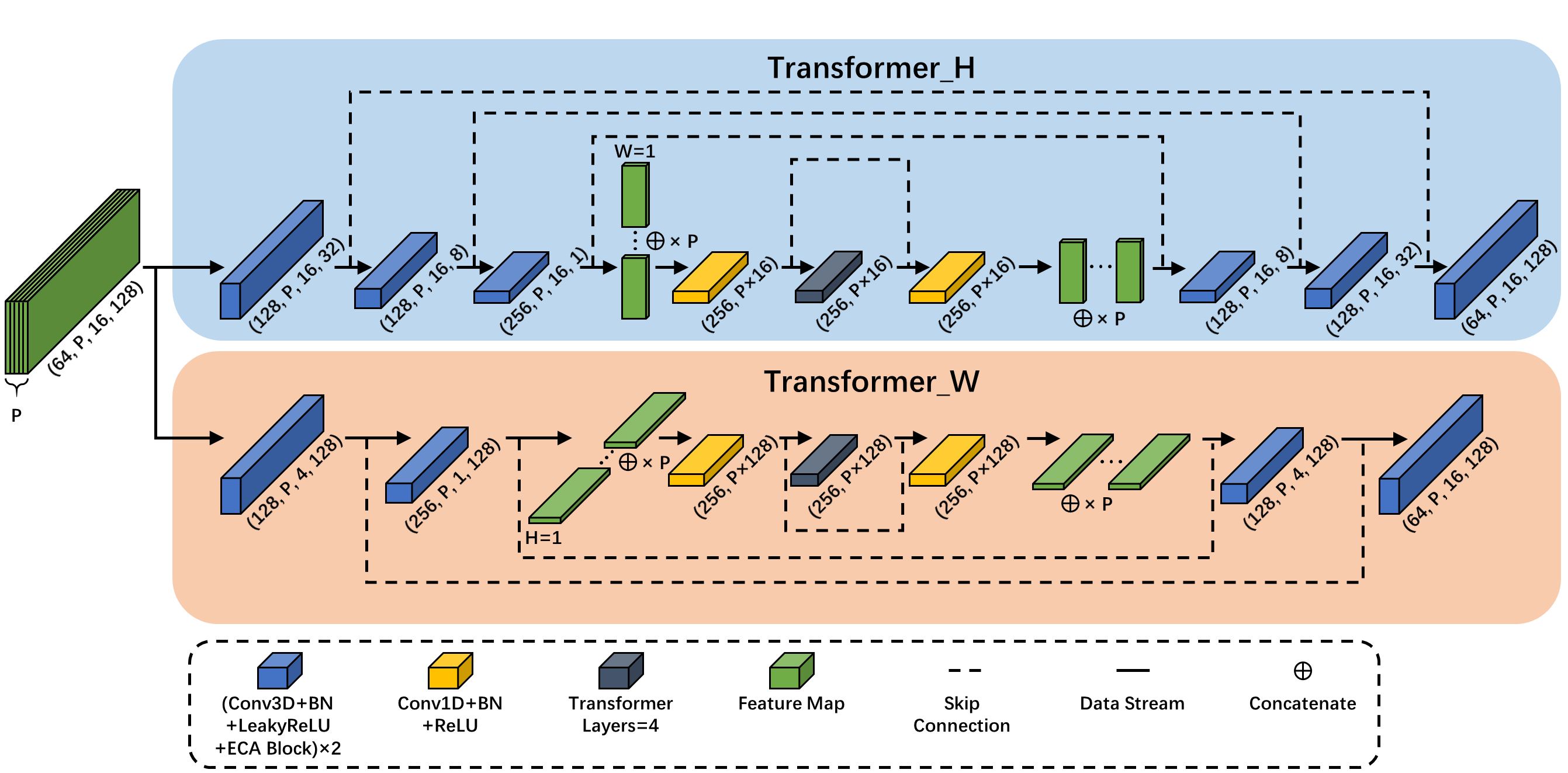}
  \caption{The architecture of Transformer block. The feature volumes from the sequential range images are concatenated into a longer image sentence before Transformer, and then re-concatenated into the previous shape after Transformer.}
  \label{fig:transformer}
  \vspace{-0.5cm}
\end{figure*}
\vspace{-0.3cm}
\subsection{Transformer Block}
\label{sec:Transformer_block}

The use of Transformer allows the network to notice the correlation between different locations throughout the input sentence~\cite{vaswani2017attention,ma2022ral}. An important problem to be solved in applying Transformer to point clouds is to generate the input features with the sentence-like form $\left( C_{l},L\right) $, where $C_{l}$ represents the number of channels and $L$ is the length of the sentence. Therefore, we design the Transformer block which is shown in~\figref{fig:transformer}. The tensors generated by the encoder are fed to the Transformer\_H block and the Transformer\_W block respectively. The down-sampling module maintains one dimension in height or width while compressing the other to the size of $1$. Then the compressed tensors are concatenated along the time dimension to obtain a continuous image sentence with a larger width. The image sentence is input into Transformer, then upsampled to the previous size through the mirror operation.

Our method is well suited for Transformer to extract spatio-temporal features. Taking the Transformer\_W block as an example, the height dimension of the feature volume is compressed to size $1$, and thus the channel dimension contains more distinct information. Each column of the image sentence aggregates all the information of a width slice. The concatenated image sentence can be expressed as $\left\lbrace w_{i} ~ \vert ~ i=1,2,...,P \cdot W_{l} \right\rbrace $, where $w_{i}$ represents the word vector at the width $i$ and $W_{l}$ represents the length of each of the $P$ image sentences. In detail, the concatenated image sentence contains the temporal position and the spatial position of each vertical slice. Therefore, Transformer can capture the spatio-temporal correlation between vertical slices in the image sentence, and further improves the performance of point cloud prediction.



\subsection{Semantic Auxiliary Training}
\label{sec:semantic}

\begin{figure}
  \centering
  \includegraphics[width=0.9\linewidth]{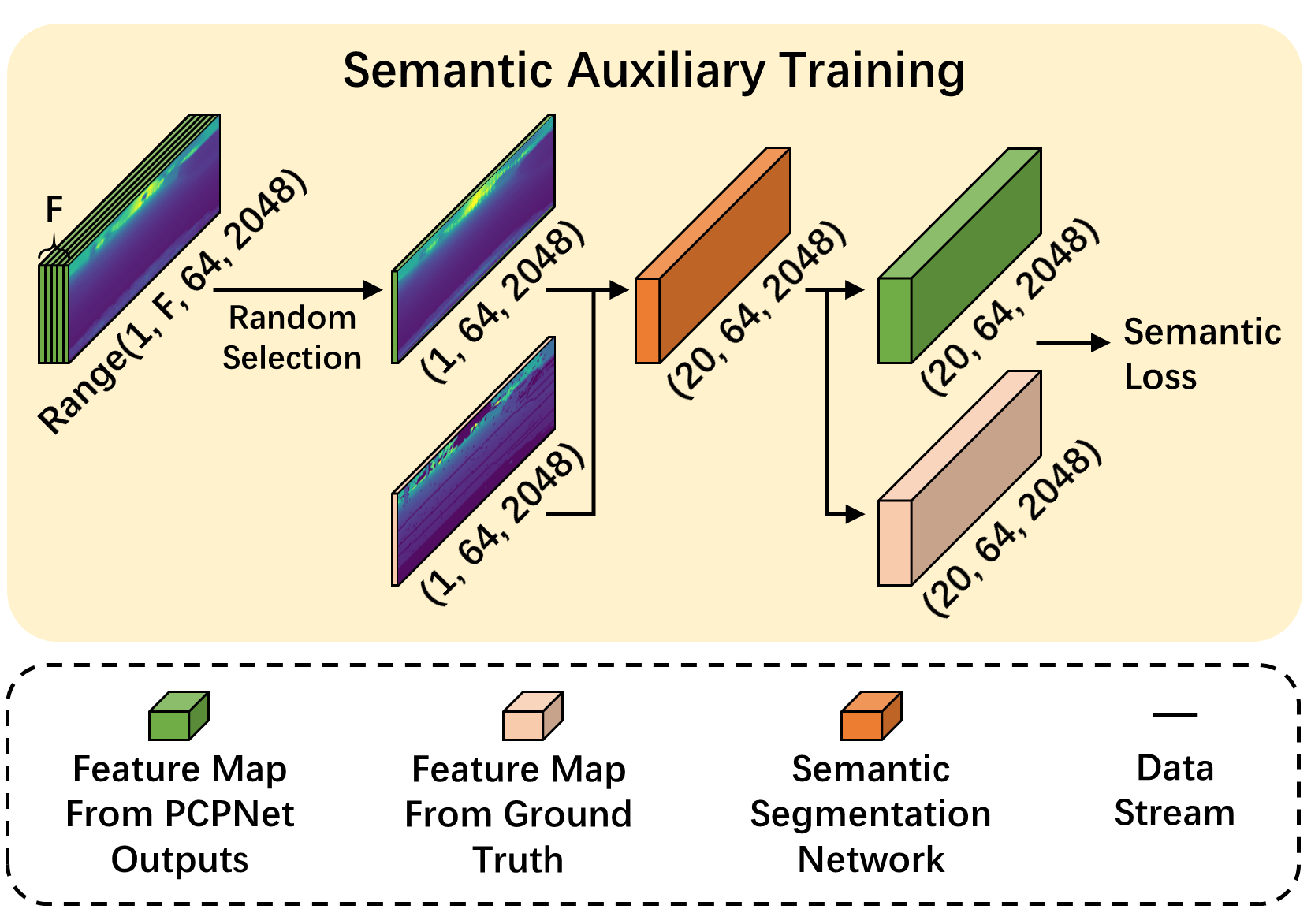}
  \caption{The details of semantic auxiliary training. The semantic loss is calculated directly on the multi-channel feature volume output by the segmentation network.}
  \label{fig:semamtic}
  \vspace{-0.5cm}
\end{figure}

\begin{figure*}
  \centering
  \includegraphics[width=1\linewidth]{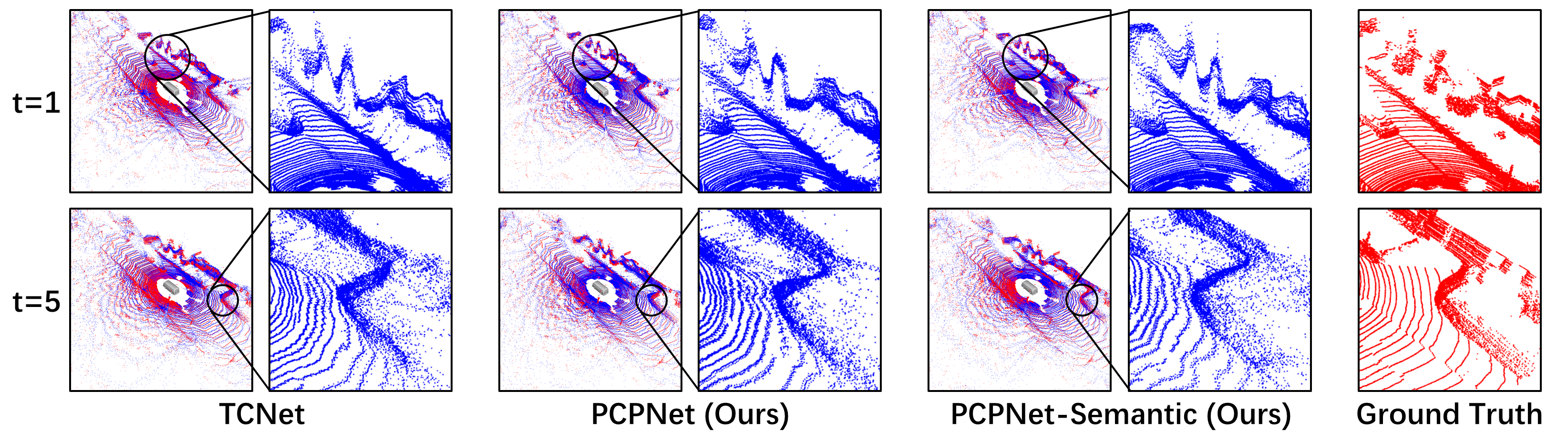}
  \caption{Qualitative comparison conducted on sequence $08$ of the KITTI dataset. The predicted points (blue) and the ground truth points (red) are combined for better visual comparison. The upper row shows the predicted step $t=1$ and the lower row the predicted step $t=5$. Local structures in large-scale point clouds are circled and enlarged to better observe local details.}
  \label{fig:qualitative}
  \vspace{-0.5cm}
\end{figure*}

In real vehicle applications, the predicted point cloud sequences can be used to serve the following downstream tasks such as future scene understanding~\cite{nunes2022segcontrast} and object detection~\cite{chen2022mppnet}. Therefore, the predicted point cloud needs to be closer to the ground truth in semantics to improve its practical value. Motivated by this, we propose semantic auxiliary training to enhance the performance of PCPNet. As shown in~\figref{fig:semamtic}, we randomly select one of the $t$ range images from the predicted sequence generated by PCPNet for one forward propagation. The selected range image is then fed to the pre-trained RangeNet++~\cite{milioto2019rangenet++}, a lightweight semantic segmentation network together with the corresponding ground truth range image. The output tensor of the segmentation network is $\left( C_{s},H_{s},W_{s}\right) $, where $C_{s}$ represents the number of classes in semantics. We calculate $L1$ loss between the semantic map from the output of PCPNet and the one from the ground truth to obtain the semantic loss. Note that the random selection in one prediction aims to reduce the calculation costs and inference time. Besides, we use the semantic output from ground truth range images as the labels to calculate the semantic loss. This helps to achieve the self-supervised training process since no manually annotated labels of semantic segmentation are needed.

We do not perform the $\argmax$ operation on channels and calculate the cross-entropy loss as most semantic segmentation methods do~\cite{qi2017pointnet, milioto2019rangenet++, liu2019meteornet} because the output of the segmentation network is probability distributions for different classes, and even the wrong semantic results contain features that are worth learning. $L1$ loss can make the semantic probability distribution between the output and the ground truth closer, rather than simply making the output close to a certain class.
\vspace{-0.5mm}
\subsection{Loss Function}
\label{sec:loss}

We use a combination of multiple losses including the average range loss $ \mathcal{L}_{R}$, the average mask loss $ \mathcal{L}_{M}$, the average semantic loss $ \mathcal{L}_{S}$, and the chamfer distance loss $ \mathcal{L}_{C}$ to train our network. The average range loss $ \mathcal{L}_{R}$ between the predicted range images $\hat{r}_{c,i,j} \in \RR^{F \times H_{f} \times W_{f}}$ and the ground truth range images $r_{c,i,j} \in \RR^{F \times H_{f} \times W_{f}}$ can be formulated as
\begin{align}
\mathcal{L}_{R} = \frac{1}{F \times H_{f} \times W_{f}} \sum\limits_{c,i,j} \Vert \hat{r}_{c,i,j}-r_{c,i,j} \Vert_{1} , \label{eq:LossR}
\end{align}
where $\Vert \bullet \Vert_{1}$ represents $L1$ norm. Since there are invalid points on the ground truth range images, we only calculate $ \mathcal{L}_{R}$ using the valid points. We further calculate the binary cross-entropy loss between the mask images $\hat{m}_{c,i,j} \in \RR^{F \times H_{f} \times W_{f}}$ and the ground truth mask images $m_{c,i,j} \in \RR^{F \times H_{f} \times W_{f}}$ to get the average mask loss $ \mathcal{L}_{M}$ by 
\begin{align}
\mathcal{L}_{M} =  \frac{1}{F \times H_{f} \times W_{f}} \sum\limits_{c,i,j} & [ -m_{c,i,j} \log{\hat{m}_{c,i,j}}  \notag
\\ &   -(1-m_{c,i,j}) \log{(1-\hat{m}_{c,i,j}) ] } , \label{eq:LossM}
\end{align}
where $\hat{m}_{c,i,j}$ is the predicted probability to demonstrate whether $(c,i,j)$ is a valid point. $m_{c,i,j}=1$ represents that the ground truth point $(c,i,j)$ is valid and $m_{c,i,j}=0$ otherwise. The average semantic loss $ \mathcal{L}_{S}$ between a predicted multi-channel semantic image $\hat{s}_{c,i,j} \in \RR^{C_{s} \times H_{s} \times W_{s}}$ which is randomly selected and the corresponding ground truth multi-channel semantic map $s_{c,i,j} \in \RR^{C_{s} \times H_{s} \times W_{s}}$ is given by
\begin{align}
 \mathcal{L}_{S} = \frac{1}{C_{s} \times H_{s} \times W_{s}} \sum\limits_{c,i,j} \Vert \hat{s}_{c,i,j}-s_{c,i,j} \Vert_{1} , \label{eq:LossS}
\end{align}
which is also calculated only at the valid points just like $ \mathcal{L}_{R}$.

Besides the $ \mathcal{L}_{R}$, $ \mathcal{L}_{M}$, and $ \mathcal{L}_{S}$ which are range-image-based losses, we also calculate the point-based loss to further improve the accuracy of point cloud prediction.
Following~\cite{fan2019pointrnn,lu2021monet,mersch2022self}, we use the Chamfer Distance~\cite{fan2017point} to measure the difference between the predicted point cloud $\hat{S} = \lbrace \hat{p_{i}} \in \RR^{3} ~\vert ~ i = 1,2,...,N \rbrace$ reprojected from the masked range image and the ground truth point cloud $S = \lbrace p_{i} \in \RR^{3} ~\vert ~ i = 1,2,...,M \rbrace$. The chamfer distance loss $ \mathcal{L}_{C}$ is calculated by
\begin{align}
 \mathcal{L}_{C} =  \frac{1}{N} \sum\limits_{\hat{p} \in \hat{S}} \min\limits_{p \in S} \Vert \hat{p}-p \Vert^{2}_{2} + \frac{1}{M} \sum\limits_{p \in S} \min\limits_{\hat{p} \in \hat{S}} \Vert \hat{p}-p \Vert^{2}_{2}, \label{eq:LossC}
\end{align}
where $\Vert \bullet \Vert_{2}$ represents $L2$ norm. Therefore, the total loss function that we ultimately use is
\begin{align}
 \mathcal{L} = \mathcal{L}_{R} + \mathcal{L}_{M} + \alpha_{S}\mathcal{L}_{S} + \alpha_{C}\mathcal{L}_{C}, \label{eq:LossAll}
\end{align}
where $\alpha_{S}$ and $\alpha_{C}$ represent the weight coefficients to demonstrate whether using $\mathcal{L}_{S}$ and $\mathcal{L}_{C}$ for training respectively.

\section{Experiments}
\label{sec:experiments}

\begin{table*}[t]
  \centering
  \setlength{\tabcolsep}{3.4mm}
  \renewcommand\arraystretch{1.1}
  \caption{Chamfer Distance Results on the KITTI Test Set}
  \footnotesize{
\begin{tabular}{c|cccccc|ccc}
\toprule
&\multicolumn{6}{c|}{Sampled Point Clouds}    &\multicolumn{3}{c}{Full-scale Point Clouds}  \\ \hline

\tabincell{c}{Prediction\\Step}  & \tabincell{c}{PointLSTM\\~\cite{fan2019pointrnn}}  & \tabincell{c}{MoNet-LSTM\\~\cite{lu2021monet}} & \tabincell{c}{MoNet-GRU\\~\cite{lu2021monet}} & \tabincell{c}{TCNet\\~\cite{mersch2022self}} & \textbf{\tabincell{c}{PCPNet \\ (Ours)}} & \textbf{\tabincell{c}{PCPNet- \\ Semantic \\ (Ours)}}  & \tabincell{c}{TCNet\\~\cite{mersch2022self}} & \textbf{\tabincell{c}{PCPNet \\ (Ours)}} & \textbf{\tabincell{c}{PCPNet- \\ Semantic \\ (Ours)}}   \\ \hline

1     & 0.317  & 0.286  & 0.278  & 0.290  & 0.285           & \textbf{0.280}  & 0.256  & 0.252           & \textbf{0.247} \\ 
2     & 0.507  & 0.412  & 0.392  & 0.357  & 0.341           & \textbf{0.340}  & 0.314  & 0.302           & \textbf{0.298} \\ 
3     & 0.750  & 0.567  & 0.543  & 0.441  & \textbf{0.411}  & 0.412           & 0.387  & 0.363           & \textbf{0.361} \\ 
4     & 0.982  & 0.719  & 0.681  & 0.522  & \textbf{0.492}  & 0.495           & 0.459  & \textbf{0.436}  & \textbf{0.436} \\ 
5     & 1.210  & 0.874  & 0.830  & 0.629  & \textbf{0.580}  & 0.601           & 0.554  & \textbf{0.514}  & 0.530          \\ \cline{1-10}
Mean  & 0.753  & 0.572  & 0.545  & 0.448  & \textbf{0.422}  & 0.426           & 0.394  & \textbf{0.373}  & 0.374          \\
\bottomrule
\end{tabular}
  }
  \label{tab:chamfer_distance}
  \vspace{-0.5cm}
\end{table*}

\subsection{Experimental settings}
\label{sec:exp_Set}

We train our proposed PCPNet in a self-supervised manner since no manually annotated labels are needed. The length of the past point cloud sequence $P$ and the future point cloud sequence $F$ are both set to $5$, which means the time step $t \in [1, 5]$. Each range image in the input and output sequences has the size $64\times2048$, and the probability threshold for each mask image is set to $0.5$ to mask out possible invalid points in the predicted range images. We use sequences $00$ to $05$ of KITTI Odometry dataset~\cite{geiger2012we} for training, sequences $06$ and $07$ for validation, and sequences $08$ to $10$ for testing. During the training process, we use the Adam optimizer~\cite{kingma2014adam} with default parameters and set the initial learning rate to $10^{-3}$ with an exponential decay weight $0.99$. Semantic auxiliary training exploits RangeNet++ which is pre-trained on SemanticKITTI~\cite{behley2019semantickitti} as the semantic segmentation network. Our proposed network and all the baseline models are trained for $50$ epochs. During the experiments, $\alpha_{S}$ is set to $1.0$ for the model using semantic auxiliary training and $0.0$ otherwise. In addition, we set $\alpha_{C}=0.0$ for pre-training and $\alpha_{C}=1.0$ for fine-tuning to realize better performance and less training time. We conduct all the following experiments on a system with an Intel i$7$-$10875$H CPU and an Nvidia RTX $3070$ GPU. 

\subsection{Qualitative Evaluation}
\label{sec:qual}

We first make a qualitative comparison between PCPNet and the range-image-based baseline method TCNet~\cite{mersch2022self} to show the superiority of our method intuitively, and the results are shown in~\figref{fig:qualitative}. PCPNet does not use semantic auxiliary training while PCPNet-Semantic does throughout the whole training process. In the predicted sequence, the $t=1$ frame shows that both PCPNet and PCPNet-Semantic outperform TCNet in terms of the structural details of the surrounding environments. From the parts enclosed and magnified by the black circles in~\figref{fig:qualitative}, the walls predicted by PCPNet and PCPNet-Semantic are closer to the ground truth. The distribution of point clouds predicted by PCPNet is more regular and less fluctuated compared to the results from TCNet. Besides, our proposed methods can also predict the shape of the car better than TCNet, and the PCPNet-Semantic forecasts best because our proposed semantic auxiliary training further helps to maintain the structure information in the semantic level.

With the increase of time steps, the prediction of point clouds becomes more difficult since there is a larger time gap between the current perception and the predicted one. At the $t=5$ frame, the prediction of TCNet at wall corners has an obvious deviation compared to the results of PCPNet. PCPNet-Semantic outperforms all the baseline methods overall, which indicates the use of semantic information enhances the ability to predict the future shape of the perceived objects with large time gaps.

\subsection{Quantitative Evaluation}
\label{sec:quan}

We quantitatively compared our methods with multiple baselines including PointLSTM~\cite{fan2019pointrnn}, MoNet-LSTM~\cite{lu2021monet}, MoNet-GRU~\cite{lu2021monet}, and TCNet~\cite{mersch2022self} on the KITTI test set, and the results support that our proposed PCPNet achieves the state-of-the-art performance on point cloud prediction. Here we use Chamfer distance $[m^{2}]$ as the metric to measure the difference between the predicted point cloud and the ground truth point cloud. PointLSTM, MoNet-LSTM, and MoNet-GRU are all point-based methods, so they use down-sampled point clouds to accelerate calculation, while TCNet and our proposed PCPNet are range-image-based methods and can predict full-scale point clouds. For the point-based methods, we follow the operation reported by their authors to downsample the input point clouds to $16384$ points to save the computing cost, while the more lightweight architectures allow the range-image-based methods PCPNet and TCNet to maintain a very low computing cost on more laser points. Since our method predicts range images, direct down-sampling for the reprojected point clouds significantly affects the final prediction results. Therefore, we follow the operation of TCNet~\cite{mersch2022self} to downsample the input point clouds to $65536$ points to compare with the point-based baseline methods. The results of the quantitative evaluation are shown in~\tabref{tab:chamfer_distance}. In terms of the sampled point clouds, our methods produce more stable predictions than other baselines as the prediction steps increase, which is reflected in the smaller chamfer distance for the larger prediction steps. Even affected by the down-sampling operation, our methods perform better on average chamfer distance at every single step. We further provide an evaluation on full-scale point clouds in~\tabref{tab:chamfer_distance}. It can be seen that our methods also perform better than the other range-image-based method TCNet at step $1\sim5$ on full-scale point clouds. In general, PCPNet-Semantic only outperforms PCPNet on chamfer distance with smaller time gaps, but can forecast the future structure information with larger time gaps which further support the clarification in~\secref{sec:qual}. 
\vspace{-0.3cm}
\subsection{Generalization Study}
\label{sec:generalization}

To prove the solid generalizability of our proposed method, we also conduct a generalization study on the nuScenes dataset~\cite{caesar2020nuscenes} with the training strategy similar to the experiments on the KITTI dataset. We use scenes $00$ to $69$ for training (70 in total), scenes $70$ and $84$ for validation (15 in total), and scenes $85$ to $99$ for testing (15 in total). Since our proposed method is self-supervised once semantic labels are available, the semantic segmentation network utilized in the auxiliary training strategy affects the generalization ability of PCPNet most. In this experiment, PCPNet is trained in a self-supervised manner on the nuScenes dataset, where the semantic labels are provided by RangeNet++ which is still pre-trained on the SemanticKITTI dataset. Since the KITTI dataset contains point clouds collected by a 64-beam LiDAR while the nuScenes dataset uses a 32-beam one, we re-train RangeNet++ on the SemanticKITTI dataset with range images of size $32\times1024$ to adapt to the input data form of nuScenes dataset. As shown in~\tabref{tab:nuScenes}, all losses of PCPNet-Semantic are less than TCNet on the nuScenes dataset, which supports that our method generalizes well into other driving environments even with semantic auxiliary training. Due to the use of a 32-beam LiDAR in the nuScenes dataset, the amount of point cloud data decreases significantly, which results in larger chamfer distance of both PCPNet and TCNet compared to the evaluation on the KITTI dataset.



\begin{table}[t]
  \centering
  \vspace{0.2cm}
  \setlength{\tabcolsep}{4mm}
  \renewcommand\arraystretch{1.1}
  \caption{Generalization Study on the nuScenes Test Set}
  \footnotesize{
\begin{tabular}{c|cccc}
\toprule
Approach     & $ \mathcal{L}_{R}$   & $ \mathcal{L}_{M}$  & $ \mathcal{L}_{S}$  & $ \mathcal{L}_{C}$  \\ \hline
TCNet    & 0.719      & 0.240    & 0.043   & 1.389      \\ \cline{1-5}
\textbf{PCPNet-Semantic}     & \textbf{0.704}       & \textbf{0.236}       & \textbf{0.034}   & \textbf{1.360}
\\ \bottomrule
\end{tabular}
  }
  \label{tab:nuScenes}
  \vspace{-0.5cm}
\end{table}

\vspace{-0.3cm}
\subsection{Ablation Study}
\label{sec:ablation}

\begin{table}[t]
  \centering
  \vspace{0.2cm}
  \setlength{\tabcolsep}{4.3mm}
  \renewcommand\arraystretch{1.1}
  \caption{Ablation Study on the KITTI Validation Set}
  \footnotesize{
\begin{tabular}{c|C{1cm}C{1cm}C{2cm}}
\toprule
Approach             & $ \mathcal{L}_{R}$   & $ \mathcal{L}_{M}$   & Runtime (ms)  \\ \hline
PCPNet-W                & 0.784                & 0.296                & \textbf{5.20}      \\ \cline{1-4}
PCPNet-H                & 0.763                & 0.293                & 6.02               \\ \cline{1-4}
\textbf{PCPNet}     & \textbf{0.742}       & \textbf{0.288}       & 8.72
\\ \bottomrule
\end{tabular}
  }
  \label{tab:ablation}
  \vspace{-0.5cm}
\end{table}

In the Transformer block of PCPNet, the input features are compressed along the height and width dimensions respectively, and then enhanced by the self-attention mechanism, which is introduced in~\secref{sec:Transformer_block}. To further validate the effectiveness of the proposed Transformer block, we conduct the ablation study on the KITTI validation set with two baselines, PCPNet-W and PCPNet-H. PCPNet-W only maintains Transformer\_W and discards Transformer\_H, while PCPNet-H only uses Transformer\_H rather than Transformer\_W. Here we use full-size point clouds as inputs and use $ \mathcal{L}_{R}$ and $ \mathcal{L}_{M}$ as the evaluation metrics. As shown in~\tabref{tab:ablation}, PCPNet outperforms PCPNet-W and PCPNet-H on both losses, which means that the two types of Transformer enhance the performance of point cloud prediction together. Besides, PCPNet-H performs better than PCPNet-W indicating that Transformer can capture more distinct spatio-temporal information along the height dimension than the width dimension. We also show the average runtime of one prediction in~\tabref{tab:ablation}. As can be seen, PCPNet costs larger inference time due to its more complex architecture than the baselines with only one Transformer block.
\vspace{-0.5mm}
\subsection{Study on Semantic Auxiliary Training}
\label{sec:valid_semantic}

In this experiment, we compared TCNet, PCPNet, and PCPNet-Semantic on the KITTI validation set to verify the enhancement from the proposed semantic auxiliary training. Here we propose a novel metric named as \textit{semantic similarity} to measure the difference between the semantic map $\hat{y}_{c,i,j} \in \RR^{C_{s} \times H_{s} \times W_{s}}$ output by the semantic segmentation network and the ground truth semantic label $y_{c,i,j} \in \RR^{C_{s} \times H_{s} \times W_{s}}$, which can be formulated as
\begin{align}
\textit{Semantic Similarity} =  \frac{C_{s} \times H_{s} \times W_{s}}{\sum\limits_{c,i,j}-y_{c,i,j} \log{\hat{y}_{c,i,j}}}. \label{eq:semantic_similarity}
\end{align}

According to~\eqref{eq:semantic_similarity}, the greater the semantic similarity, the closer $\hat{y}_{c,i,j}$ is to $y_{c,i,j}$. Note that our method is completely self-supervised and we can only use the semantic map of the ground truth point clouds to train the network, but here we evaluate the effectiveness of the semantic auxiliary training using annotated semantic labels from SemanticKITTI dataset. The comparison of the semantic similarity on the KITTI validation set is shown in~\tabref{tab:semantic_similarity}, where Ground Truth represents the semantic similarity between the semantic map from the ground truth range image and the ground truth semantic label, and is regarded as the upper bound performance. As can be seen, our methods outperform TCNet in the semantic level. Besides, the semantic similarity of PCPNet-Semantic is further improved by applying semantic auxiliary training.

\begin{table}[t]
  \centering
  \vspace{0.2cm}
  \setlength{\tabcolsep}{5.5mm}
  \renewcommand\arraystretch{1.1}
  \caption{Comparison of Semantic Similarity on the KITTI Validation Set}
  \footnotesize{
\begin{tabular}{c|C{3cm}}
\toprule
Approach                          & Semantic Similarity     \\ \hline
TCNet                             & 2.789                   \\ \cline{1-2}
PCPNet (Ours)                & 2.876                   \\ \cline{1-2}
\textbf{PCPNet-Semantic (Ours)}   & \textbf{2.913}          \\ \cline{1-2}
Ground Truth (Upper Bound)                     & 3.877                   \\ \bottomrule
\end{tabular}
  }
  \label{tab:semantic_similarity}
  \vspace{-0.5cm}
\end{table}

\begin{figure}
\vspace{0.2cm}
  \centering
  \includegraphics[width=0.9\linewidth]{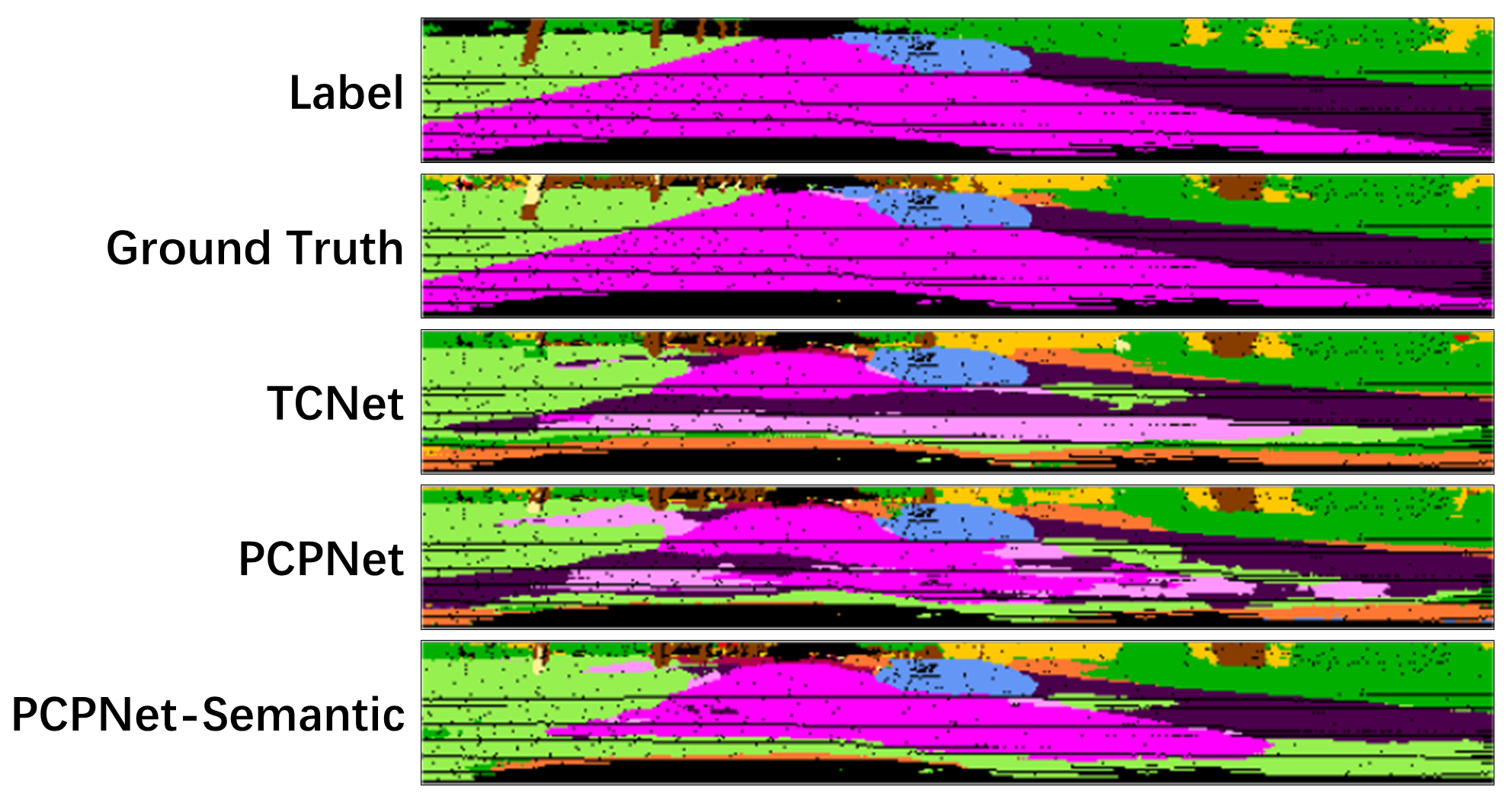}
  \caption{Visualization of the outputs of semantic segmentation network. \textit{Label} refers to the manual labels from SemanticKITTI, and \textit{Ground Truth} refers to the semantic map obtained from the ground truth point clouds.}
  \label{fig:valid_semamtic}
  \vspace{-0.5cm}
\end{figure}

\begin{table}[t]
  \centering
  \vspace{0.2cm}
  \setlength{\tabcolsep}{1.7mm}
  \renewcommand\arraystretch{1.1}
  \caption{Comparison of Losses on the KITTI Validation Set}
  \footnotesize{
\begin{tabular}{c|C{1cm}C{1cm}C{1cm}C{1cm}}
\toprule
Approach                         & $ \mathcal{L}_{R}$   & $ \mathcal{L}_{M}$   & $ \mathcal{L}_{S}$  & $ \mathcal{L}_{C}$  \\ \hline
TCNet                            & 0.8143               & 0.2990               & 0.0445              & 0.4796  \\ \cline{1-5}
PCPNet (Ours)               & 0.7865               & 0.2932               & 0.0423              & 0.4541 \\ \cline{1-5}
\textbf{PCPNet-Semantic (Ours)}  & \textbf{0.7694}      & \textbf{0.2914}      & \textbf{0.0324}     & \textbf{0.4510}
\\ \bottomrule
\end{tabular}
  }
  \label{tab:loss_semantic}
  \vspace{-0.5cm}
\end{table}

The quantitative experimental results in~\tabref{tab:loss_semantic} show that all losses of PCPNet-Semantic decrease, especially the average semantic loss $ \mathcal{L}_{S}$ which is $23.4\%$ lower than PCPNet. In addition, our approach still performs better than TCNet even without the enhancement from semantic auxiliary training.

The visualization of the comparison results is shown in~\figref{fig:valid_semamtic}. Both TCNet and PCPNet lose some semantic information, such as roads and grasses. In contrast, this situation is alleviated by PCPNet-Semantic and more laser points are correctly classified. 
\vspace{-1mm}
\subsection{Complexity Analysis}
\label{sec:complexity}


\begin{table}[t]
  \centering
  \vspace{0.2cm}
  \setlength{\tabcolsep}{5.15mm}
  \renewcommand\arraystretch{1.1}
  \caption{Complexity Analysis on the KITTI Test Set}
  \footnotesize{
\begin{tabular}{cc|cc}
\toprule
\multicolumn{2}{c|}{Approach}   & \tabincell{c}{FLOPs\\(billion)} & \tabincell{c}{Params\\(million)} \\ \hline
\multicolumn{1}{r|}{\multirow{2}{*}{PointLSTM}}    & 16384\,pts       & 50.01  &  1.23          \\ \cline{2-4}
\multicolumn{1}{c|}{}                              & 65536\,pts      & 200.04   & 1.23            \\ \cline{1-4}       
\multicolumn{1}{r|}{\multirow{2}{*}{MoNet-LSTM}}   & 16384\,pts      & 92.76  & 4.00     \\ \cline{2-4}
\multicolumn{1}{c|}{}                              & 65536\,pts       & 371.03 & 4.00 \\ \cline{1-4}
\multicolumn{1}{r|}{\multirow{2}{*}{MoNet-GRU}}    & 16384\,pts         & 59.76 & 3.31 \\ \cline{2-4}
\multicolumn{1}{c|}{}                              & 65536\,pts        & 239.06  & 3.31 \\ \cline{1-4}
\multicolumn{2}{c|}{TCNet}                         & 30.26  & 17.01    \\ \cline{1-4}
\multicolumn{2}{c|}{PCPNet (Ours)}                   & 54.34 & 22.60
\\ \bottomrule
\end{tabular}
  }
  \label{tab:complexity}
  \vspace{-0.5cm}
\end{table}

\begin{figure}[t]
\vspace{0.2cm}
  \centering
  \includegraphics[width=0.85\linewidth]{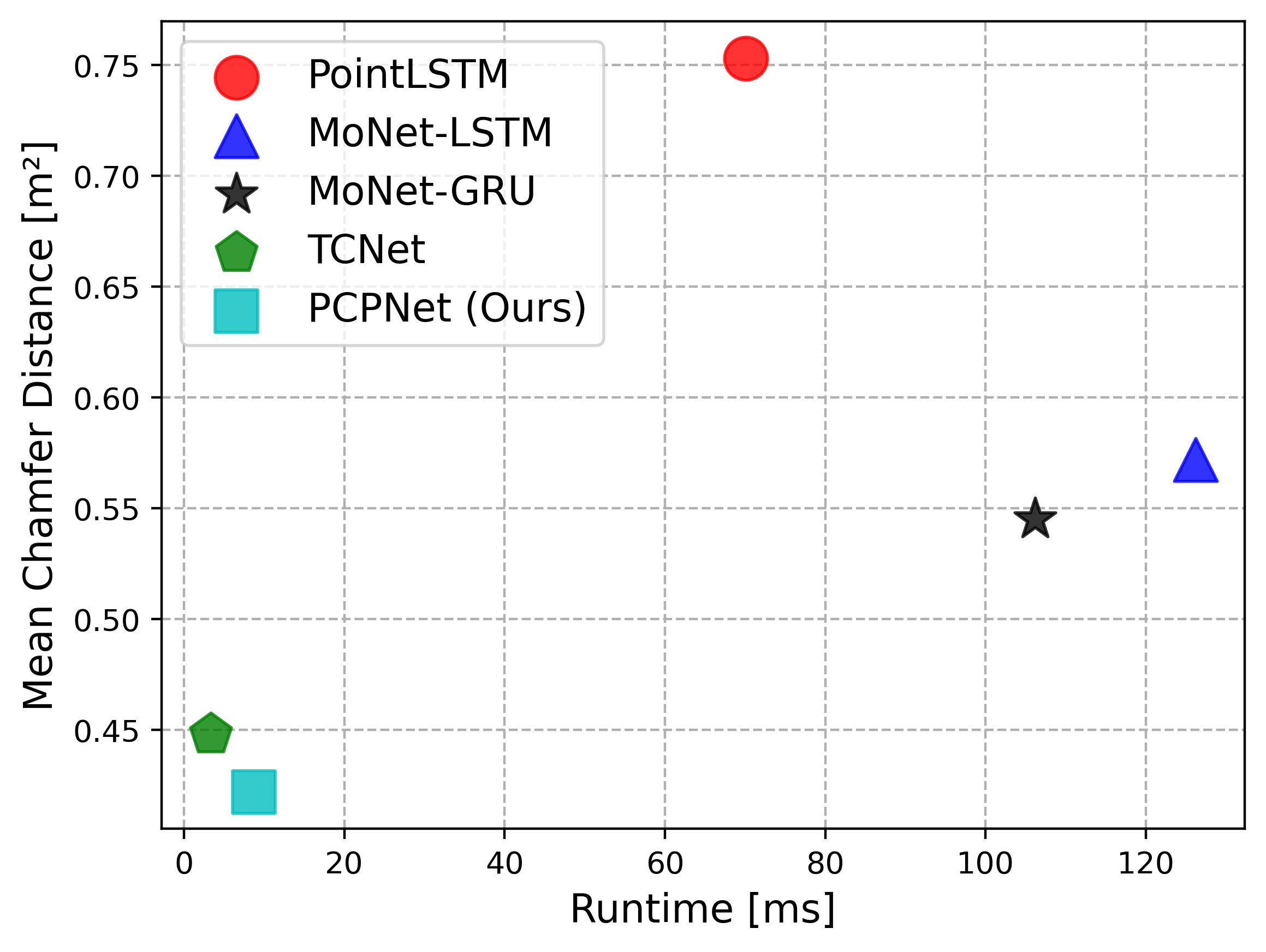}
  \caption{The performance on chamfer distance v.s. runtime on the KITTI test set. PointLSTM, MoNet LSTM, and MoNet GRU use 16384 downsampled points as inputs to improve inference speed.}
  \label{fig:complexity_analysis}
  \vspace{-0.5cm}
\end{figure}

In this experiment, we first compare the runtime of PCPNet with other baseline methods. The performance on chamfer distance v.s. runtime of networks on the KITTI test set is illustrated in~\figref{fig:complexity_analysis}. As can be seen, our proposed PCPNet performs best on chamfer distance while maintaining good real-time performance ($8.72$\,ms to predict $5$ future point clouds). Moreover, the complexity analysis of all the PCP methods is shown in~\tabref{tab:complexity}. Compared with the point-based methods, the range-image-based methods have relatively lower time complexity. For example, the FLOPs of the MoNet-GRU is $239.06$ billion for predicting $65536$ points, while the FLOPs of our proposed PCPNet is only $54.34$ billion with points predicted twice as MoNet-GRU. Although the time and space complexities of PCPNet are slightly greater than the other range-image-based method TCNet, PCPNet has better performance in predicting future point clouds due to the sophisticated network architecture.

\section{Conclusion}
\label{sec:conclusion}

In this paper, we propose a self-supervised method to predict future point cloud sequences based on the given past point clouds. Benefiting from the self-attention mechanism of Transformer, our proposed network can aggregate spatio-temporal information along multiple dimensions. We also propose a semantic auxiliary training strategy to enhance the performance of forecasting more realistic point clouds for real-vehicle applications. The proposed network is evaluated on publicly available datasets using multiple metrics, and the experimental results support that our method outperforms the other state-of-the-art methods in point cloud prediction while maintaining a very fast running speed.

In the future, more types of semantic loss functions except for $L1$ norm can be adopted for an ablation study in different driving scenes. Furthermore, it may be possible to discuss whether our proposed semantic auxiliary training strategies can improve the performance of other point cloud prediction models in the future.


\bibliographystyle{unsrt}

\footnotesize{
\bibliography{glorified,new}}

\end{document}